# Performance Assessment of ChatGPT vs Bard in Detecting Alzheimer's Dementia

Dr. Balamurali B T (PhD) and Dr. Jer-Ming Chen (PhD)

Science, Mathematics & Technology (SMT), Singapore University of Technology & Design
8 Somapah Rd, 487372, Singapore.

Email of corresponding authors: balamurali_bt@sutd.edu.sg & jerming_chen@sutd.edu.sg

**Abstract**

Large language models (LLMs) find increasing applications in many fields. Here, three LLM chatbots (ChatGPT-3.5, ChatGPT-4 and Bard) are assessed – in their current form, as publicly available – for their ability to recognize Alzheimer's Dementia (AD) and Cognitively Normal (CN) individuals using textual input derived from spontaneous speech recordings. Zero-shot learning approach is used at two levels of independent queries, with the second query (chain-of-thought prompting) eliciting more detailed than the first. Each LLM chatbot's performance is evaluated on the prediction generated in terms of accuracy, sensitivity, specificity, precision and F1 score. LLM chatbots generated three-class outcome ("AD", "CN", or "Unsure"). When positively identifying AD, Bard produced highest true-positives (89% recall) and highest F1 score (71%), but tended to misidentify CN as AD, with high confidence (low "Unsure" rates); for positively identifying CN, GPT-4 resulted in the highest true-negatives at 56% and highest F1 score (62%), adopting a diplomatic stance (moderate "Unsure" rates). Overall, three LLM chatbots identify AD vs CN surpassing chance-levels but do not currently satisfy clinical application.

# 1. Introduction

Large Language Model (LLM) chatbots such as OpenAI's Generative Pretrained Transformer (GPT, versions 3.5 and 4) and Google's Bard, demonstrate impressive capabilities in many domains including healthcare settings [1–3] to support early detection and clinical assessment. Here, we explore the utility of LLM chatbots (ChatGPT-3.5, ChatGPT-4 and Bard) for identifying Alzheimer's Dementia (AD) in individuals, using textual transcriptions derived from spontaneous speech: a non-trivial assessment task that currently poses significant challenges for other state-of-the-art detection modalities and could benefit immediately from advanced artificial intelligence techniques directly applied "in-the-field".

In our recent work [4], we adopted a multi-modal approach that integrated audio- and text-based techniques to automatically detect AD from audio recordings of spontaneous speech. Notably, our text-based classification method exhibited superior performance (88.7% accuracy) compared to audio-based classification (78.9% accuracy). At the time of reporting, this state-of-the-art accuracy surpasses other contemporaneous studies utilizing the same dataset [5–8], highlighting the efficacy of text-based approach in reliably identifying AD. Thus motivated, we investigate here the potential viability of LLM chatbots – in their 'as-is' form (i.e. utilizing zero-shot learning in which the model is expected to make predictions about classes it has not been explicitly trained on [9,10]) – as a possible tool to detect AD using only spontaneous speech. While previous studies using audio, text, or their fusion with model embeddings as features achieved promising AD/CN classification accuracies ranging from 78-88% [4–8,11–14], they all relied on supervised learning procedure with labelled training data. This limits their generalizability to real-world settings lacking readily available labels. In contrast, the zero-shot LLM chatbot learning approach in the current investigation assesses the feasibility of using chatbots in-the-field without requiring labelled training data. Therefore, direct comparisons of LLM chatbots' performance with existing results from supervised models would be inappropriate. Instead, our investigation focuses on evaluating the appropriateness and clinical efficacy of LLM chatbot responses as a supplementary framework for AD assessment.

## 2. Methods

We survey three state-of-the-art LLM chatbots: OpenAI's language models GPT-3.5 & GPT-4 (14 March 2023 ChatGPT version; http://openai.com) [15] and Google's language model, Bard (10 May 2023 version; https://bard.google.com) [16]. Throughout this paper, 'GPT-3.5' and 'GPT-4' refer to OpenAI's corresponding ChatGPT 3.5 and ChatGPT 4 generative pre-trained transformer chatbots. To investigate their ability to identify subjects with AD compared to Cognitively Normal (CN) subjects based on the input text, we employed a zero-shot learning approach, where LLM chatbots were presented with the transcribed text as a single input at two levels of independent prompts, with the second query more detailed (a less structured chain-of-thought prompting approach, a strategy which is different from step-by-step thinking [10]) than the first:

- Query 1 (Q1). "Could the following transcribed speech be from a Cognitive Normal or Alzheimer's Dementia subject?"
- Query 2 (Q2). "Can you look at the syntax, vocabulary, structure, narration style, grammar, semantic discourse, stylistics, pragmatics and share your opinion in short concise points on what you think of the following paragraphs? These paragraphs are transcribed text from an interview with different subjects. Could they be narrated by a Cognitive Normal or an Alzheimer's Dementia subject?"

A well-studied dataset of audio recordings provided in the ADReSSo Challenge [17] was utilized in this study. Specifically, these audio recordings captured interview sessions where participants described the "Cookie Theft Picture" from the Boston Diagnostic Aphasia Examination [18] in English. Audio segments containing the interviewer's speech, including any overlap with the subject, were removed, while retaining non-speech segments such as silence and filler words. To facilitate textual input, the speech audio was transcribed into text using Otter.ai platform [19]. Given the zero-shot nature of our approach, we focused on the 71 recordings (36 CN, 35 AD) constituting the testing set of ADReSSo Challenge [17], aligning with our prior work [4] for comparison and generalizability assessment.

Text from each recording, accompanied with the queries (Q1 and Q2) were presented as input prompts only once to each LLM chatbot. The corresponding output responses generated by each LLM chatbot, with typical phrases such as "This paragraph appears to be narrated by someone with AD," "This narration style could be from a CN individual" or "The paragraph could be narrated by both a CN or an AD subject", were then classified accordingly as "AD", "CN", or "Unsure" categories. The relationship between predicted LLM chatbots outcomes for each recording and subjects' Mini-Mental State Examination (MMSE) scores was examined (cf. Figure 3) to identify any specific trends or patterns.

A note on repeatability: Although LLM chatbots are known to evolve over time as they dynamically learn and update their knowledge continuously, our preliminary investigation on repeatability revealed that the classification outcomes made within a week remained largely consistent (within 10% consistent). This implies a level of stability in the predictions made by LLM chatbots, at least within the investigative timeframe (GPTs: week of 13 March 2023; Bard: week of 10 May 2023). (Upon querying the temperature parameter (the parameter that controls the randomness and creativity, a higher temperature results in creative, diverse responses, but results in factual error, while a lower temperature creates conservative, less-engaging responses and are usually accurate) used for prompt replies, ChatGPT reported a fixed value set around 0.7. In contrast, Bard's temperature setting is characterized by its dynamic nature, adapting to the specific requirements of each prompt and context [20].).

## 3. Results

The results of this investigation are divided into four sections: (1) outcome matrices (True vs Predicted classes) for Query 1 and Query 2 are summarized and compared for three LLMs; (2) performance metrics of each LLM targeting AD and CN classes are then scored separately. (3) Analyzing Chain-of-thought prompting (Query 2) offering insight into linguistic motivations, (4) Comparing LLMs' prediction outcome against the subject's Mini-Mental State Examination (MMSE) scores.

### 3.1 Performance & Outcomes from Prompts (Q1 & Q2)

In response to prompts Q1 and Q2, the output generated by all three LLM chatbots fell into three categories: AD, CN, and an additional category of "Unsure" (see confusion matrices in Figure 1)*. The appearance of the "Unsure" category indicates that in some instances LLM chatbots faced challenges in classifying AD vs CN confidently and equivocated on the cognitive status of certain subjects. This uncertainty arises unsurprisingly (in hindsight), as it reflects the complexity and variability of linguistic patterns in spontaneous speech, as well as limitations of zero-shot learning approach. It also signals agency on the LLM chatbots' part to offer a third option despite a binary choice clearly solicited.

- *For a three-class outcome, naively considering zero-shot outcomes arise randomly with equal probability, chance level is arguably 33%. While our zero-shot queries yielded three-class outcomes (as would be typically encountered) despite eliciting binary ground truths, expecting the two-class outcome at 50% chance level (typical in supervised learning settings) becomes inappropriate: firstly, it disregards the full spectrum of nuanced responses captured by chatbots in this zero-shot context; secondly, it introduces bias by forcing data into a pre-defined, binary framework, which does not capture the chatbot's true behaviour.

| | True Class | Q1 AD | Q1 Unsure | Q1 CN | Q2 AD | Q2 Unsure | Q2 CN |
|---|---|---|---|---|---|---|---|
| GPT-3.5 | AD | 60.0% | 17.1% | 22.9% | 82.9% | 8.6% | 8.6% |
| | CN | 55.6% | 11.1% | 33.3% | 58.3% | 8.3% | 33.3% |
| GPT-4 | AD | 11.4% | 57.1% | 31.4% | 28.6% | 51.4% | 20.0% |
| | CN | 2.8% | 38.9% | 58.3% | 19.4% | 27.8% | 52.8% |
| Bard | AD | 88.6% | 2.9% | 8.6% | 88.6% | 11.4% | 0% |
| | CN | 58.3% | 13.9% | 27.8% | 63.9% | 25.0% | 11.1% |

**Predicted Class**

**Fig 1**: Summary matrices (True vs Predicted) of three LLM chatbots studied for Query 1 (Q1) and Query 2 (Q2), showing cognitive classification outcomes for three class prediction (AD *vs* CN *vs* Unsure) and their occurrence rate (%) when presented with the same text dataset from AD and CN subjects (True Class, 35 and 36 subjects, respectively).

GPT-3.5 demonstrated good performance in correctly detecting AD subjects, with accuracy 60% and 83% for prompts Q1 and Q2, respectively. However, its performance in correctly identifying CN subjects was only moderate, with 33% accuracy (chance level) for both prompts. Notably, GPT-3.5 exhibited a pronounced tendency to misclassify CN subjects as AD, with high misidentification rates 56% and 58% for the two prompts, respectively. GPT-3.5 generally displayed confidence in its predictions, as indicated by the relatively low rates of "Unsure" responses (17% and 9% for AD subjects; 11% and 8% for CN subjects, across the two prompts respectively).

GPT-4, on the other hand, excelled in correctly detecting CN subjects, achieving accuracies 58% and 53% for Q1 and Q2 prompts, respectively. However, its performance in correctly identifying AD subjects was poor, with accuracies 11% and 29% for the two prompts. GPT-4 displayed a moderate tendency to misidentify AD subjects as CN, with misidentification rates 31% and 20% for the two prompts, respectively. Notably, the model exhibited a ‘diplomatic’ response pattern where it equivocated with the highest rates of “Unsure” responses (57% and 51% for AD subjects; 39% and 28% for CN subjects, across the two prompts respectively), hesitant to commit to a prediction between AD and CN, while showing greater uncertainty for AD subjects than CN subjects.

Bard demonstrated the strongest performance in correctly detecting AD subjects, achieving 89% accuracy for both Q1 and Q2 prompts. However, it was the poorest among the three LLM chatbots at identifying CN subjects correctly, with accuracies 28% and 11% for the two prompts, respectively. Bard also displayed a strong tendency to misclassify CN subjects as AD, with misidentification rates 58% and 64% for the prompts, respectively. Like GPT-3.5 (and unlike GPT-4), Bard exhibited confidence in its predictions, as reflected by low rates of “Unsure” responses (2% and 11% for AD subjects; 14% and 25% for CN subjects, across the two prompts respectively).

The three LLM chatbots show fair consistency regarding Q1 and Q2 prompts, indicating at a gross level that the models’ performance is not sensitive to how the query is structured per se – a straightforward query suffices if no further supporting details or insight is sought. Both Bard and GPT-3.5 perform similarly (and better) at correctly detecting AD subjects compared to GPT-4; in contrast, GPT-4 performed better at identifying CN subjects, while also preferring ‘diplomatic’ responses contributing to “Unsure” prediction class.

## 3.2 Performance Metrics of Three LLM chatbots for Q1 & Q2

Traditional binary classification metrics, such as sensitivity, specificity, and precision, are readily computable but become problematic for our investigation: all three LLM chatbots queried yield three potential outputs (AD, CN, and "Unsure") despite the binary classification problem (AD vs. CN). Several approaches could address this mismatched output response: (1) Forced choice prompts – LLMs could be prompted to explicitly yield AD or CN outputs, even if the LLM is uncertain. This can be achieved by incorporating additional instructions within the prompts. (2) Binning "Unsure" as AD – considering all "unsure" outputs as AD (reflecting the clinical practice that "Unsure" cases are flagged for further investigation). In this study, however, we chose to analyze the performance of LLM chatbots separately when predicting for AD class and when predicting for CN class, as we recognize these two prediction tasks are neither complementary nor symmetric; accordingly, this nuanced approach provides better insight into the respective strengths/weaknesses, biases, and transparency of each LLM chatbot.

Specifically, when focusing on predicting AD (true class), the complement duly consists of the predicted "Unsure" and CN classes. Conversely, when focusing on predicting CN (true class), the complement accordingly comprises the "Unsure" and AD classes. Table 1 summarizes the performance metrics (accuracy, sensitivity, specificity, precision, and F1 score) derived for the three LLM chatbots in response to Q1 and Q2 prompts in Figure 1.

**Table 1:** Performance metrics of the three LLM chatbots for Query 1 and Query 2, focusing on predicting AD (top) and predicting CN (bottom) respectively. Highest performances averaged across both queries are shaded in orange.

| **AD Predicted** | GPT-3.5 | | GPT-4 | | Bard | | Metrics Averaged (across respective LLM) |
|---|---|---|---|---|---|---|---|
| Query | Query 1 | Query 2 | Query 1 | Query 2 | Query 1 | Query 2 | |
| Accuracy | 0.46 | 0.58 | 0.35 | 0.41 | 0.58 | 0.49 | (0.52+0.38+0.54)/3 = 0.48 |
| Sensitivity | 0.60 | 0.83 | 0.11 | 0.29 | 0.89 | 0.89 | (0.72+0.20+0.89)/3 = 0.60 |
| Specificity | 0.38 | 0.36 | 0.95 | 0.73 | 0.32 | 0.15 | (0.37+0.84+0.24)/3 = 0.48 |
| Precision | 0.51 | 0.58 | 0.80 | 0.59 | 0.60 | 0.57 | (0.55+0.70+0.59)/3 = 0.61 |
| F1 Score | 0.55 | 0.68 | 0.20 | 0.38 | 0.71 | 0.70 | (0.62+0.29+0.71)/3 = 0.54 |
| Overall Mean | 0.50 | 0.61 | 0.48 | 0.48 | 0.62 | 0.56 | (0.56+0.48+0.59)/3 = 0.54 |

| **CN Predicted** | GPT-3.5 | | GPT-4 | | Bard | | Metrics Averaged (across respective LLM) |
|---|---|---|---|---|---|---|---|
| Query | Query 1 | Query 2 | Query 1 | Query 2 | Query 1 | Query 2 | |
| Accuracy | 0.46 | 0.58 | 0.35 | 0.41 | 0.58 | 0.49 | (0.52+0.38+0.54)/3 = 0.48 |
| Sensitivity | 0.33 | 0.33 | 0.58 | 0.53 | 0.28 | 0.11 | (0.33+0.56+0.20)/3 = 0.36 |
| Specificity | 0.72 | 0.91 | 0.27 | 0.59 | 0.91 | 1.00 | (0.82+0.43+0.96)/3 = 0.73 |
| Precision | 0.60 | 0.80 | 0.66 | 0.73 | 0.77 | 1.00 | (0.70+0.70+0.89)/3 = 0.76 |
| F1 Score | 0.43 | 0.47 | 0.62 | 0.61 | 0.41 | 0.20 | (0.45+0.62+0.31)/3 = 0.46 |
| Overall Mean | 0.51 | 0.62 | 0.50 | 0.57 | 0.59 | 0.56 | (0.57+0.54+0.58)/3 = 0.56 |

Comparing Q1 and Q2 prompts reveal that, on average, Q2 elicits better performance for most metrics in GPT-3.5 and GPT-4, but not for Bard.

When comparing the performance of predicting AD versus predicting CN, the aggregated overall mean of the performance metrics of the three LLM chatbots for the two queries together suggests that predicting CN (56%) is slightly better than predicting AD (54%); both tasks exhibit relatively comparable performance, regardless of model. GPT-3.5 and Bard perform similarly well in predicting both CN and AD, achieving 57% and 58% for predicting CN, and 56% and 59% for predicting AD, respectively on average. On the other hand, GPT-4 performs relatively better at predicting CN (average of 54%) compared to predicting AD (average of 48%). Notably, both GPT-3.5 and Bard outperform GPT-4 overall.

Depending on the task, certain performance metrics can be exceptionally high. For instance, when predicting CN, Bard achieves 100% specificity and precision. Similarly, when predicting AD, GPT-4 achieves 95% specificity, while Bard achieves 89% sensitivity. Additionally, predicting CN tends to exhibit more extreme performance metrics ranging from 36% to 76%, compared to predicting AD, which ranges from 48% to 61%.

Because of the presence of the “Unsure” class arising, performance metrics of the three LLM chatbots vary depending on whether the focus is to identify AD subjects or to identify CN subjects, except for the accuracy metric (identical for both objectives).

In terms of overall accuracy (considering the average of both Q1+Q2 outcomes together), GPT-3.5 and Bard again performed comparably, with scores averaging 52% and 54% respectively. However, GPT-4 had the lowest overall accuracy at 38% (just above chance level*). Bard exhibited the highest sensitivity (89%) when predicting AD (averaged for Q1+Q2), while GPT-4 showed the highest sensitivity (56%) when predicting CN. GPT-4 demonstrated the highest specificity (84%) and precision (70%) when predicting AD (averaged for Q1+Q2), while Bard achieved the highest specificity (96%) and precision (89%) when predicting CN.

F1 score, which indicates how precision and sensitivity balance out at the expense of the other (with the proviso that both measures are equally important), reveals that Bard robustly outperforms the other LLM chatbots in predicting AD (achieved the highest F1 score of 71%, averaged for Q1+Q2); GPT-4 on the other hand shows the lowest F1 score at 29% (averaged for Q1+Q2). Interestingly, the reverse occurs when predicting CN: GPT-4 achieved the highest F1 score of 62% (averaged for Q1+Q2), while Bard showed the lowest F1 score at 31% (averaged for Q1+Q2).

These findings underscore the understanding that predicting for AD and predicting for CN are disparate objectives, with challenges which are not simply complementary nor commensurate. Accordingly, the LLM chatbots respond differently in their capacity to correctly distinguish between AD and CN. However, from a healthcare standpoint, it is essential to highlight that LLM chatbots-based approach capable of achieving sensitivity exceeding 58% – as demonstrated by GPT-4 for Q1 (the most robust among all query and LLM chatbots) for classifying CNs into a control group – holds promise. This emphasis is vital, considering approaches that misdiagnose a notable portion of healthy individuals as having dementia, potentially result in distressing false-positive outcomes.

### 3.3 Insights from Chain-of-Thought Prompting (Query 2)

Query 2 solicits the LLM chatbots' chain-of-thought and reveals intermediate insights (linguistic features such as syntax, vocabulary, structure, narration style, grammar, semantic discourse, stylistics, and pragmatics) that drive its classification decisions. Accordingly, Figure 2 depicts the Query 2 output responses – aggregated and visualized as word clouds – for two AD and two CN subjects who were consistently classified correctly by all three LLM chatbots. These word clouds provide a visual representation of the prominent linguistic attributes and patterns identified by LLM chatbots, in terms of word (semantic) classes and frequency of occurrence in the LLM chatbots' output response.

Word clouds associated with AD predictions demonstrate a greater visual spread of descriptor words and are characterized by linguistic attributes such as "incoherence", "disorganization", "fragmented", and "disjointed". These findings align with known linguistic observations describing AD [21], where subjects often exhibit difficulty maintaining coherence and producing organized speech. In contrast, CN word clouds appear visually sparser, focusing heavily on two or three main keywords, and are associated with attributes such as "coherent", "straightforward", and "organized". This suggests that LLM chatbots are likely relying on higher-order linguistic features to differentiate between AD and CN subjects. Figure 2 visually summarizes the distinct linguistic training and patterns utilized by each LLM for the classification task.

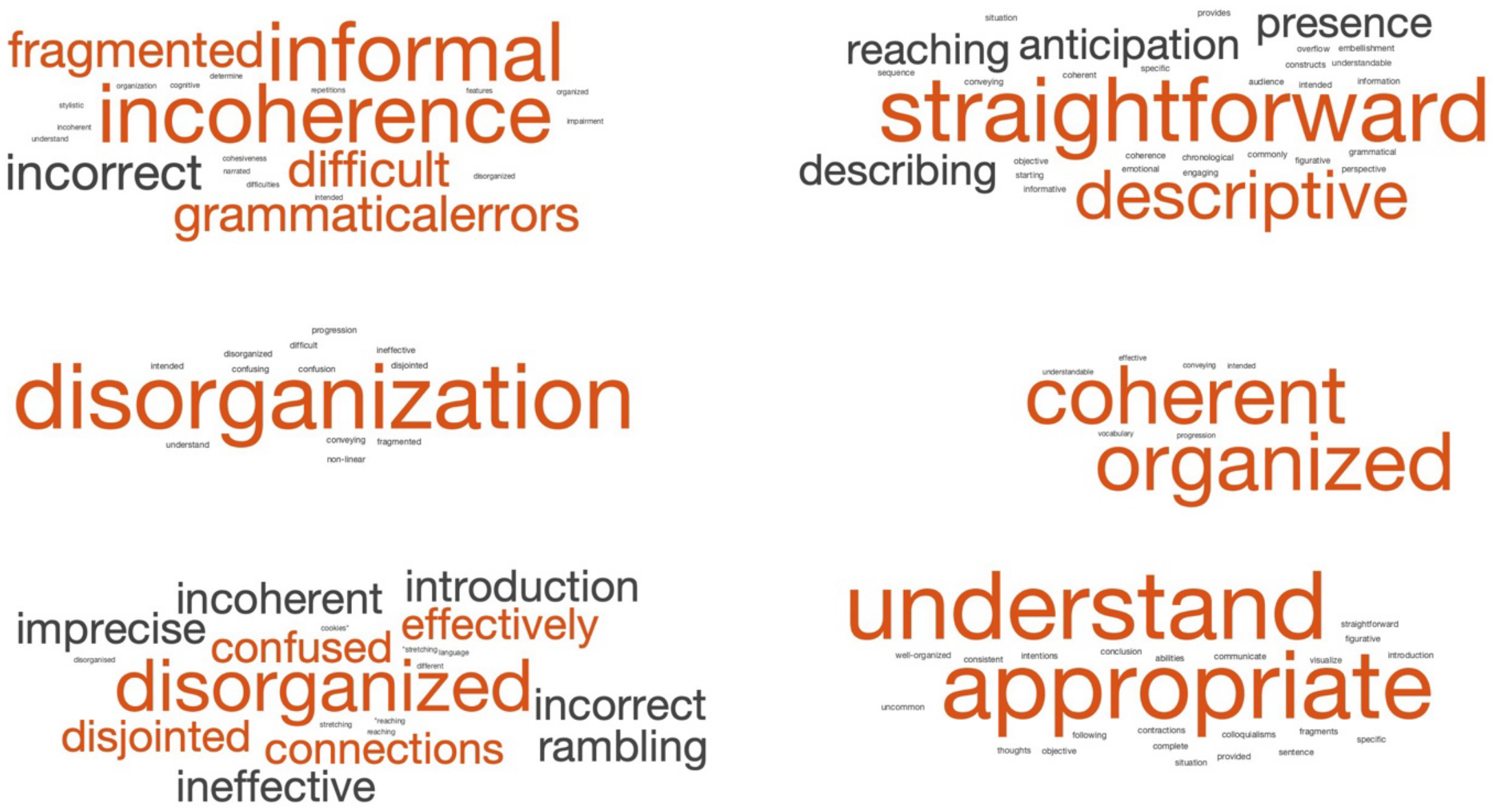

**Fig 2.** Word Cloud summaries for correct AD prediction (left) and correct CN prediction (right) text output generated by GPT-3.5 (top), GPT-4 (middle) and Bard (bottom), for two AD subjects and two CN subjects.

## 3.4 Insights from MMSE Score Comparison

The Mini–Mental State Examination (MMSE) is a common measure of cognitive impairment, where the maximum score is 30 (normal cognitive function) while scores below 23 or 24 indicate possible cognitive decline [22,23], though sociocultural variables such age and education could affect individual scores [24]. Accordingly, Figure 3 plots MMSE scores for all subjects against the prediction response (AD/Unsure/CN) classed by the three LLM chatbots, alongside the score distribution for all AD and CN subjects. Note: two AD subjects possess MMSE scores exceeding 25 (and one CN subject scored at 24), suggesting a degree of heterogeneity within both groups.

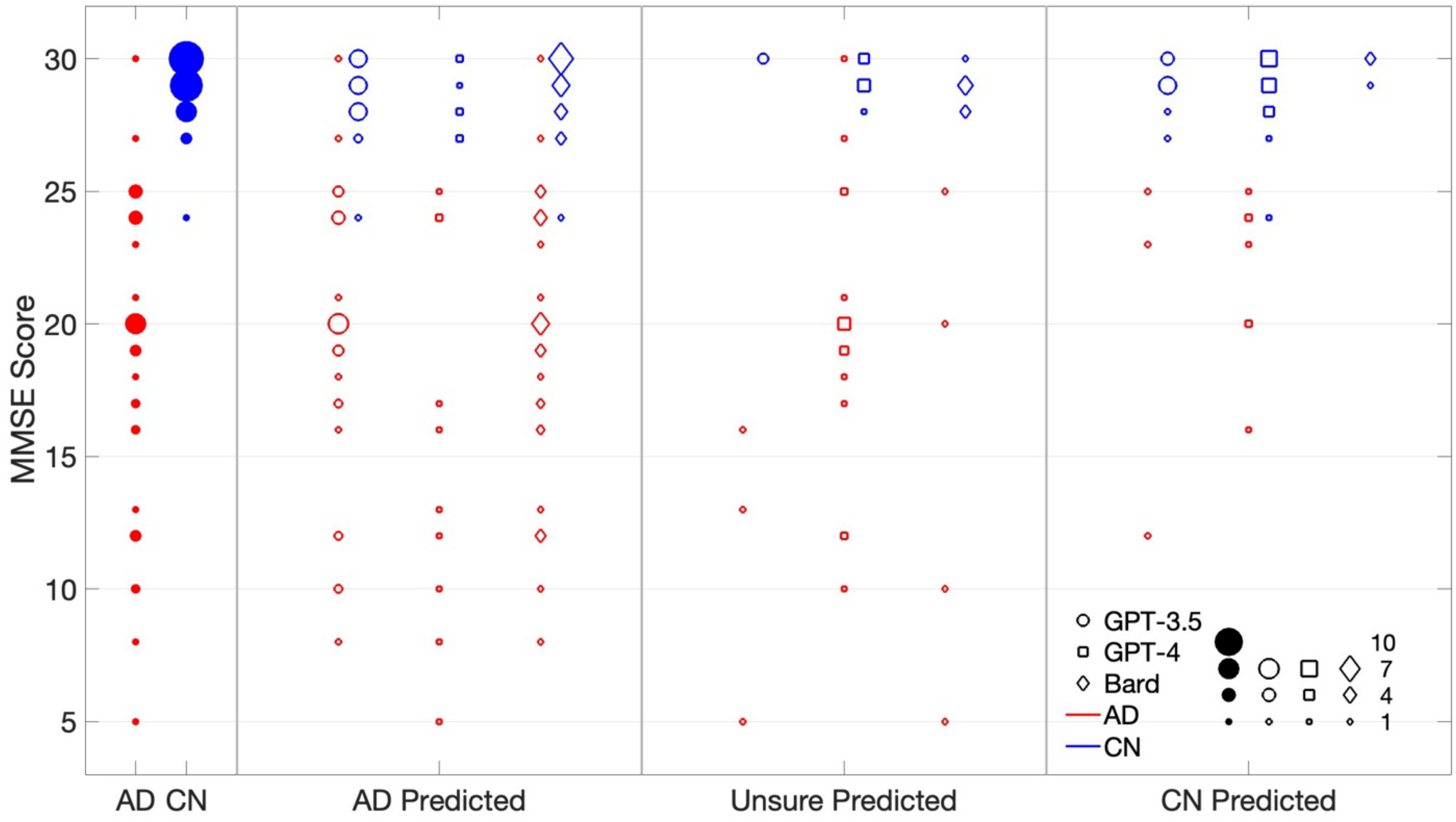

**Fig 3:** MMSE Score vs Prediction Classes (AD/CN/Unsure) of Q2 for AD (red) and CN (blue) subjects across three LLM chatbots (circle: GPT-3.5; square: GPT-4; diamond: Bard); the leftmost filled circles depict MMSE score distribution for all AD and CN subjects (True Class). The size of the symbols indicates the relative frequency of occurrence of that MMSE value (legend, bottom right).

Regardless of the prediction class (AD/Unsure/CN), there does not appear to be a clear relationship between MMSE scores and the LLM chatbots' prediction performance for CN subjects (high MMSE scores are inherently limited in distribution). However, a significant subset of CN subjects with high MMSE scores (27-30) tend to be misclassified as AD or Unsure by all three LLM chatbots, suggesting that high MMSE scores does not necessarily aid CN prediction; other factors, such as linguistic attributes or contextual cues, may offer a confounding impact on the LLM chatbots' prediction.

Among AD subjects, GPT-3.5 demonstrates a broad range of MMSE scores (8-30) for correctly classified AD subjects yet predicts "Unsure" even for subjects with lower scores (5-16). Additionally, it occasionally misclassifies AD subjects as CN for slightly higher MMSE scores (12-25), which may be deemed reasonable. Similarly, GPT-4 exhibits a wide distribution of MMSE scores (5-25) for correctly classified AD subjects, with scores tending to be slightly lower than those associated with “Unsure” predictions (10-30) and CN predictions (16-25). In contrast, Bard avoids predicting CN for AD subjects altogether but displays a diverse range of MMSE scores (8-30) for accurate AD classification, a trend rather like GPT-3.5’s prediction (including frequency). Notably, all three models demonstrate substantial variation in MMSE scores for correctly classified AD subjects, highlighting the tenuous relationship between MMSE scores and LLM chatbot predictions, and the limits of its utility to enhance prediction accuracy.

While the absence of a low MMSE score may lead to misclassifications of AD subjects as CN, the opposite does not hold true – high-scoring AD subjects are still correctly classified by the LLM chatbots. Overall, there does not appear to be a strong correlation between MMSE score and prediction performance for either AD or CN subjects. This is not unexpected, as cognitive acuity of an individual varies at different instances across time, whereas the subject’s speech-based performance assessed here on a single picture-description task represents only a snapshot in time.

## 4. Discussion

Our study assessed the use of LLM chatbots for distinguishing between AD and CN classifications based on text transcriptions and provided insight into the challenges of each task and the suitability of each LLM chatbot for that objective. For positively identifying CN, GPT-4 emerges as the preferred choice, surpassing chance-level performance (true-negative at 56%). However, it should be noted that GPT-4 tends to adopt a diplomatic stance without committing to a clear prediction between AD and CN. Bard, on the other hand, stands out for positively identifying AD with 88.6% true-positive rate. However, a limitation is observed in its tendency to misidentify CN as AD, often with high confidence. In terms of overall performance metrics, both GPT-3.5 and Bard demonstrate comparable performance in positively identifying CN.

Existing approaches for AD and CN classification employ audio, text, or their fusion, achieving accuracies of 78.9% (audio alone), 84.5% (text alone), and a range of 80.2 to 88.7% with various deep neural network models applied to fusion strategies [4–8,11–14]. Notably, these reported models rely on supervised learning with labelled training data. In contrast, the performance of LLM chatbots in this study does not match the level achieved by the aforementioned supervised models. However, making direct comparisons between the performance of LLM chatbots and the reported supervised models would be inappropriate due to significant differences in their underlying learning paradigms.

Why does predicting AD result in more 'homogenous' performance (*cf.* Table 1) than predicting CN? First, because subjects with AD typically exhibit distinct linguistic impairments, including incoherence, disorganized speech, fragmented language, and disjointed narratives, these attributes associated with AD are often more distinguishable, leading to greater consistency and identifiability for the LLM chatbots in recognizing and predicting AD. In contrast, CN subjects span a spectrum of linguistic styles, resulting in subtler differences in language use and a narrower range of standard observable linguistic markers compared to AD. These factors contribute to the complexity of predicting CN, as the wide variability in language patterns and the absence of clear markers make accurate CN identification challenging. Further, a majority of LLM training datasets likely consist of functional texts reflecting standard linguistic features drawing upon a large feature space encompassing broad narrative, persuasive, expository, and descriptive examples; these contrasts sharply with deviations from the feature space associated with linguistic impairments arising from AD (likely also drawing on more modest datasets). Two obfuscating outcomes arise accordingly: training data imbalance, and consequently, identifying CN becomes less straightforward than identifying AD.

Among the three LLM chatbots considered, Bard exhibits a notable advantage over the GPTs in predicting AD: Bard's responses exhibit a higher degree of detail for Q2, because it systematically considers sentence-level, paragraph-level, text-level, and discourse-level breakdowns. This richer granularity, combined with its ability to identify deviations in the trained feature space, likely contributes to Bard's superior AD detection capabilities.

In the context of zero-shot learning, where models are not specifically fine-tuned or primed for the classification task, the direct impact of articulating the chain-of-thought response on accurate AD vs CN classification may be expected to be limited, as zero-shot models rely on general language patterns learned during pre-training and lack explicit knowledge of task-specific labels. In our investigation, we nevertheless observed that articulating chain-of-thought response had indeed a positive effect on the overall performance of GPT models (but did not benefit Bard).

This study has several limitations:

1. Efficacy and Limitations of Prompts: Only two prompts (Q1 and Q2) were investigated here. It is worth considering whether more sophisticated prompting approaches would yield different outcomes. To necessitates the integration of such chatbots into healthcare practice further refinement through the formulation of more specific queries that can provide a deeper understanding of common language impairments is essential [25].

2. "Snapshot" & Limited Probing: The potential variations of outcomes across different accounts, machines, and repetitions at different times were not explored – although we noted during preliminary explorations of repeated prompting using the same text for a particular subject does elicit slight variations in textual response (the extent, consistency or variation of these differences was not studied nor quantified) but not enough to influence the overall prediction outcome.

3. Non-repeatability & Dynamic Evolution of LLM chatbots: The efficacy and evolution of the LLM chatbots' "personality" resulting from continuous querying and intervention ('fine-tuning') of service operators, including back-end updates to new versions, remain unknown and warrant further exploration: performance differences between GPT-3.5 and GPT-4 is a stark example of this concern.

4. Accuracy of Transcription (Source Text): An automated speech-to-text service was used to transcribe interview audio recordings, so it is expected transcriptions errors and confusion will arise when the speaker's voice is not clear (signal:noise concerns) or when the target speaker does not enunciate with clarity or speak with non-standard accent. (Note: this difficulty is faced equally by all investigators using the same dataset.)

5. De-contextualized Speech: To ensure the query used only speech originating from the subject in question (and not the interviewer), prompts from the interviewer were removed before speech-to-text transcription. Consequently, semantic content of the transcriptions may appear fragmented or discontinuous due to the missing contextual information and may influence the LLM chatbots' performance in predicting AD and CN.

6. Due to the accessibility of the Dementia Bank repository, the possibility of the training data for the LLM chatbots containing instances from ADReSSo dataset cannot be entirely ruled out. This could limit the generalizability of the findings.

7. Furthermore, it is important to delve into the longitudinal progression of speech samples, as this aspect holds the potential in alerting healthcare professionals to serve as an early indicator of cognitive decline. To facilitate this investigation, the acquisition of a language dataset comprising speech samples from individuals presenting with mild cognitive impairment (MCI) alongside knowledge of their outcomes after three years becomes essential, the timing is also important and could help in predicting progression of MCI to AD [26].

## 5. Conclusion

The three LLM chatbots surveyed demonstrate ability to identify AD vs CN surpassing chance-level performance, albeit with varying degrees of accuracy and confidence: when positively identifying AD, Bard performed best with 89% true-positive rate but tended to misidentify CN as AD, often with high confidence (low "Unsure" rates); when positively identifying CN, GPT-4 performed best in identifying this true-negative at 56% but tended to adopt a more diplomatic stance (moderate "Unsure" rates). By leveraging on the unique strengths of different LLM chatbots (in their current form, as available publicly), we evaluated the performance and suitability as a first level tool to screen for cognitive decline based on spontaneous speech. However, further refinement is still needed to ensure reliability and effectiveness of these models in real-world healthcare contexts.

**Article Information**

**Acknowledgments:** The Authors are grateful to Prof Lu Wei (SUTD) for comments and suggestions on the initial explorations in this study.
**Funding:** No funding was received for this study.
**Author contributions:**
Conceptualization: BBT, JMC
Methodology: BBT, JMC
Investigation: BBT
Visualization: BBT
Supervision: JMC
Writing – original draft: BBT, JMC
**Competing interests:** Authors declare that they have no competing interests.
**Data and materials availability:** The transcription data can be made available upon request from corresponding authors (balamurali_bt@sutd.edu.sg, jerming_chen@sutd.edu.sg).
**Supplementary Materials:** NA
**Ethics Declaration:** No human or animal subjects were included in this investigation.